\documentclass[10pt,twocolumn,letterpaper]{article}

\usepackage{cvpr} 
\usepackage{times}
\usepackage{epsfig}
\usepackage{graphicx}
\usepackage{amsmath}
\usepackage{amssymb}

\usepackage{booktabs}
\usepackage{multirow}
\usepackage{rotating}
\usepackage{makecell}
\usepackage{hyperref}
\usepackage{color}
\usepackage{enumitem}

\begin{document}

\title{Benchmarking Face Recognition without Real Faces}


\author{Paweł Borsukiewicz\\
University of Luxembourg\\
Luxembourg\\
{\tt\small pawel.borsukiewicz@uni.lu}
\and
Daniele Lunghi\thanks{Corresponding author}\\
University of Luxembourg\\
Luxembourg\\
{\tt\small daniele.lunghi@uni.lu}
\and
Wendkûuni C. Ouédraogo\\
University of Luxembourg\\
Luxembourg\\
{\tt\small wendkuuni.ouedraogo@uni.lu}
\and
Jacques Klein\\
University of Luxembourg\\
Luxembourg\\
{\tt\small jacques.klein@uni.lu}
\and
Tegawendé F. Bissyandé\\
University of Luxembourg\\
Luxembourg\\
{\tt\small tegawende.bissyande@uni.lu}
}

\maketitle
\thispagestyle{empty}

\begin{abstract}

Synthetic face datasets have become effective enough to train face recognition models with accuracy rivaling that of models trained on real photographs. 
This progress sidesteps the ethical and legal burdens of collecting real biometric data, yet evaluation has not kept pace. 
Even studies that train entirely on synthetic images still rely on real-face benchmarks to measure performance, leaving the privacy problem only half solved.

We ask whether synthetic datasets can replace real benchmarks for face recognition evaluation. 
We test 12 synthetic datasets against 7 established real benchmarks using 24 pre-trained models that span both convolutional and transformer architectures. 
Our evaluation covers biometric verification metrics, similarity score distributions, cross-model ranking consistency, and the underlying distributional properties of each dataset. 
Benchmarking fidelity varies widely across the synthetic candidates, but the two strongest, MorphFace and Vec2Face, reproduce the relative behavior of real benchmarks and reach agreement levels that fall within the natural disagreement already observed among the real benchmarks themselves.

These results establish that well-constructed synthetic datasets can support reliable comparative evaluation for face recognition, moving the field closer to a fully synthetic and privacy-preserving pipeline for both training and benchmarking.
\end{abstract}

\section{Introduction}

Biometric identifiers such as fingerprints, facial geometry, and vein patterns constitute sensitive personal data. 
Their collection and processing fall under strict legal scrutiny, including the EU's General Data Protection Regulation (GDPR)~\cite{GDPR2016a} and the Illinois Biometric Information Privacy Act (BIPA)~\cite{BIPA}. 
Beyond legal compliance, ethical obligations further constrain the biometric data life cycle. 
Data acquisition typically requires explicit informed consent and clearance from an institutional ethics committee~\cite{boutros2023synthetic}. 
These constraints limit both development and reproducibility, because many biometric datasets cannot be redistributed once privacy or consent requirements change.

Synthetic data offers a more ethical and privacy-friendly alternative~\cite{borsukiewicz2025beyond, nisevic2025synthetic, sun2025ensuring} to real biometrics, since no sensitive information is collected from real subjects and regulatory exposure is correspondingly reduced. 
Advances in generative models, from Generative Adversarial Networks (GANs)~\cite{goodfellow2014generative} to Diffusion Models (DMs)~\cite{ho2020denoising}, have made synthetic training data increasingly competitive. 
Models trained entirely on artificial faces now achieve recognition rates comparable to those of models trained on real photographs~\cite{borsukiewicz2025beyond}. 
Yet this is only half a solution. 
While the training side of the pipeline can now operate without real faces, evaluation still cannot. Even studies that propose new synthetic datasets evaluate them on real benchmarks. 
A proper assessment of whether synthetic data suffices for evaluation is, to the best of our knowledge, absent from the literature. 
Without this evidence, researchers remain forced to include real data in their pipelines, if only for benchmarking, and the privacy problem persists.

Research into synthetic benchmark datasets has been limited~\cite{baltsou2024sdfd, colbois2021use, nzalasse2024sig} and has not produced clear guidance on their practical applicability. 
In this work we provide a systematic answer to the question: \textit{can synthetic benchmarks reliably reflect model performance on real data?} This question is not merely academic. With limited control over the generation process~\cite{melzi2023gandiffface}, synthetic data can exhibit poor identity separability, insufficient intra-class diversity, and demographic biases. 
A distributional mismatch between synthetic and real images therefore cannot be excluded without thorough empirical evaluation.

The main contributions of this work are as follows:
\begin{itemize}
    \setlength{\itemsep}{0pt}
    \item We conduct the first comprehensive evaluation of 12 synthetic face recognition datasets as benchmarks across 24 face recognition models, comparing them with 7 commonly used real benchmark datasets.
    \item We provide a detailed breakdown of each dataset's benchmarking performance across recognition accuracy and a set of standard biometric evaluation metrics.
    \item We show that MorphFace and Vec2Face achieve the strongest alignment with real benchmarks and can serve as reliable substitutes.
    \item We investigate the factors that influence benchmarking utility, including domain gap, identity separability, and intra-class diversity.
    \item To promote reproducible research, our code is publicly available\footnote{\url{https://github.com/Pabilito/Synthetic-Facial-Recognition-Benchmark}}.
\end{itemize}
\section{Related Work}

Face recognition benchmarking has historically relied on real-world data (\ref{sec:real_datasets}). The growing body of synthetic datasets (\ref{sec:synt_datasets}) has recently opened the door to synthetic evaluation (\ref{sec:synt_benchmarks}).

\subsection{Real Benchmark Datasets}
\label{sec:real_datasets}

Labeled Faces in the Wild (LFW)~\cite{huang2008lfw}, with 13{,}233 images of 5{,}749 identities, was long the standard unconstrained benchmark for face recognition. 
As model performance on LFW has saturated~\cite{whitelam2017iarpa}, however, with most recent systems approaching 100\% accuracy, its ability to discriminate among models has diminished.

Several more challenging alternatives have since been introduced. Cross-Pose LFW (CPLFW)~\cite{zheng2018cplfw} and Cross-Age LFW (CALFW)~\cite{zheng2017calfw} modify the original LFW pairing strategy. 
CPLFW increases pose variability, while CALFW narrows the age gap between mated and non-mated pairs so that age alone is less predictive. 
On LFW, the mean age difference is 4.94 years for mated pairs and 14.85 years for non-mated pairs; CALFW brings these to 16.61 and 16.14, respectively. 
AgeDB-30~\cite{moschoglou2017agedb} targets temporal variation more directly by selecting 16{,}488 images from 568 identities, each pair separated by approximately 30 years. 
Celebrities in Frontal-Profile (CFP-FP)~\cite{sengupta2016cfp_fp} instead focuses on viewpoint, pairing frontal and profile images across 7{,}000 images of 500 identities.

Representativeness is an equally important concern. 
IJB-B~\cite{whitelam2017iarpa} addresses this with 1{,}845 human-labeled identities captured under diverse acquisition conditions; its 21{,}798 still images and 55{,}026 video frames carry, among others, annotations for occlusion, skin tone, facial landmarks, and localization. 
IJB-C~\cite{maze2018iarpa} extends IJB-B to 31{,}334 images of 3{,}531 identities, broadening occupational and racial diversity and introducing stronger occlusion variation.

\subsection{Synthetic Training Datasets}
\label{sec:synt_datasets}

Retractions of real face datasets~\cite{boutros2023synthetic} and potential legal consequences~\cite{smith2022ethical} have motivated the search for alternatives. Synthetic data emerged as a response to the mass-scraping practices behind many widely used datasets. The development of Generative Adversarial Networks~\cite{goodfellow2014generative} and Diffusion Models~\cite{ho2020denoising} made high-fidelity image generation feasible, and research on synthetic face datasets has accelerated accordingly. State-of-the-art generators~\cite{karras2021alias, rombach2022high} can be conditioned~\cite{boutros2023synthetic} to preserve a specific identity across multiple images.

The central difficulty~\cite{borsukiewicz2025beyond} in constructing a useful synthetic dataset is achieving high intra-class diversity, the ability to produce semi-hard~\cite{sun2024cemiface} samples that challenge the model, without sacrificing identity separability, meaning that distinct identities must not overlap unnaturally in the embedding space. Early datasets such as Syn-1M~\cite{kortylewski2018training} and SynMulti-PIE~\cite{colbois2021use} struggled with this balance and yielded lower performance when used for training. More recent datasets have closed the gap considerably. MorphFace~\cite{mi2025morphFace}, CemiFace~\cite{sun2024cemiface}, Vec2Face~\cite{wu2024vec2face}, and VariFace~\cite{yeung2024variface} now support training pipelines whose recognition performance matches, and in some cases exceeds~\cite{yeung2024variface}, that of models trained on real data.

\subsection{Synthetic Benchmark Datasets}
\label{sec:synt_benchmarks}

Only a handful of datasets have been proposed specifically for benchmarking. SynMulti-PIE~\cite{colbois2021use} was the first to address this goal, emphasizing the balance~\cite{borsukiewicz2025beyond} between identity separability (minimal overlap between the similarity distributions of genuine and impostor pairs) and intra-class diversity (sufficient variation to make evaluation challenging). The authors argue that generated samples preserve identity without revealing the real identities used to train the generator. However, their evaluation covers only 64 identities, which limits the robustness of the conclusions, particularly with respect to universality.

ControlFace10k~\cite{nzalasse2024sig} contains 10{,}008 images of 3{,}336 synthetic identities generated through the Synthetic Identity Generation (SIG) pipeline. 
Although it provides three images per identity (frontal, left profile, and right profile), the intra-class variation was reported to be insufficient~\cite{borsukiewicz2025beyond}. 
Identity separability is also weak, making ControlFace10k a questionable choice for either training or benchmarking.

SDFD~\cite{baltsou2024sdfd} consists of 1{,}000 high-quality diffusion-generated samples with one image per identity. 
It is useful for demographic attribute prediction but cannot support pairwise verification benchmarking since it lacks multiple images per identity.

No synthetic dataset, whether designed for training or evaluation, has been directly validated as a face recognition benchmark. Because synthetic images may differ from real images in distributional properties and generative biases, their suitability as evaluation tools cannot be assumed. 
Establishing this suitability through rigorous empirical comparison is the gap that our study fills.
\section{Methodology}
Our testbed comprises 7 real and 12 synthetic face recognition datasets evaluated across 24 pre-trained models.

\subsection{Datasets}
We selected 7 real datasets for evaluation: LFW, CPLFW, CALFW, AgeDB-30, CFP-FP, IJB-B, and IJB-C. Beyond being the most commonly used benchmarks in the literature, these datasets collectively cover a broad range of challenging scenarios including pose variation and age gaps.

For the synthetic side, we selected 12 datasets generated by methods ranging from pure GANs and pure diffusion-based approaches to hybrid pipelines combining DMs with 3D morphable models (see Table~\ref{tab:datasets}). 
Together, these 12 datasets constitute the majority of publicly available synthetic face data. Access to IDiff-Face and SFace2 was obtained through online request forms; access to MorphFace was obtained by contacting the authors directly.

\begin{table}[ht]
\begin{center}
\caption{Overview of synthetic datasets used.}
\vspace{2mm}
\label{tab:datasets}
\setlength{\tabcolsep}{4pt}
\small
\begin{tabular}{lccc}
\toprule
\textbf{Dataset} & \textbf{\makecell{Generation \\ Method}} & \textbf{\makecell{First \\ Release}} & \textbf{Venue} \\
\midrule
SynMulti-PIE~\cite{colbois2021use}   & GAN            & 2021 & IJCB'21    \\
SynFace~\cite{qiu2021synface}                & GAN            & 2021 & ICCV'21    \\
DCFace~\cite{kim2023dcface}                  & Diffusion      & 2022 & CVPR'23    \\
IDiff-Face~\cite{boutros2023idiff}           & Diffusion      & 2023 & ICCV'23    \\
SFace2~\cite{boutros2024sface2}              & GAN            & 2024 & T-BIOM     \\
Langevin-DisCo~\cite{geissbuhler2024Langevin}& GAN            & 2024 & ICML'25    \\
Vec2Face~\cite{wu2024vec2face}               & GAN            & 2024 & ICLR'25    \\
CemiFace~\cite{sun2024cemiface}              & Diffusion      & 2024 & NeurIPS'24 \\
ControlFace10k~\cite{nzalasse2024sig}        & Diffusion      & 2024 & ICPR'25    \\
HyperFace~\cite{shahreza2024hyperface}       & GAN            & 2024 & ICLR'25    \\
MorphFace~\cite{mi2025morphFace}             & Diffusion + 3D & 2025 & CVPR'25    \\
Digi2Real~\cite{george2025digi2real} & Diffusion + 3D & 2025 & WACV'25    \\
\bottomrule
\end{tabular}
\end{center}
\end{table}

\subsection{Models}
We selected 24 pre-trained models spanning 9 vision transformers (ViTs)~\cite{dan2023transface, george2024edgeface, qin2023swinface} and 15 convolutional neural networks (CNNs)~\cite{behrmann2019invertible, deng2019arcface, SzegedyEtAl2015}, trained on diverse datasets~\cite{Glint, VggFace2, MS1MV3, yi2014casia}. 
Because this paper focuses on synthetic data, we specifically included 7 CNNs trained entirely on synthetic data~\cite{boutros2023idiff, boutros2022sface, geissbuhler2024Langevin, shahreza2024hyperface, wu2024vec2face}, all using the ResNet50 backbone, which is the standard architecture for benchmarking synthetic face recognition. 
Table~\ref{tab:models} lists all models used.


\begin{table}[h]
\small
    \centering
    \caption{Overview of models used.}
    \vspace{2mm}
    \label{tab:models}
    \begin{tabular}{ll}
        \toprule
        \textbf{Model} & \textbf{Training dataset} \\
        \midrule
        ResNet50 
            &  Langevin-DisCo/Dispersion,\\
            & IDiff-Face, HSFace10/300k\\
            & HyperFace, SFace\\
        IResNet (18/50/100)   & Glint360K, MS1MV2 \\
        FaceNet      & VGGFace2, CASIA-WebFace \\
        SwinFace     & MS1MV2 \\
        TransFace (S, L)  & Glint360K, MS1MV2 \\
        EdgeFace (base, S, XS, XXS) & MS1MV3 \\
        \bottomrule
    \end{tabular}
\end{table}

\subsection{Evaluation Metrics}
Face recognition systems typically report recognition accuracy, the percentage of correctly classified pairs, as a primary metric. 
Accuracy alone is insufficient for real-world biometric systems, however, because the costs of false positives and false negatives differ substantially~\cite{grother2021face}. 
Following ISO/IEC 19795-1:2021~\cite{ISO19795}, security-critical systems prioritize keeping the False Match Rate (FMR), also called False Acceptance Rate (FAR), below a fixed threshold, and then minimize the False Non-Match Rate (FNMR), the proportion of genuine samples incorrectly rejected.

The acceptable FMR is application-dependent. 
Systems protecting sensitive resources demand near-zero false positives, while consumer devices tolerate a less strict decision threshold. 
This distinction is captured by FMR100 and FMR1000, which report the FNMR when the decision threshold is set so that only 1 in 100 or 1 in 1{,}000 impostor comparisons results in a false match. ZeroFMR is the FNMR at FMR${}={}$0. 
The Equal Error Rate (EER) is the operating point where FMR equals FNMR, balancing security against usability. 
Finally, TAR@FAR reports the percentage of genuine samples accepted at a given FAR threshold, complementing the FMR-based metrics.

\subsection{Dataset Correlation}
A synthetic dataset can serve as a benchmark only if the scores it produces correlate with those obtained on real data. Given $n$ paired observations ($x_i$, $y_i$) representing the same models evaluated on two different datasets, we measure linear agreement with Pearson's coefficient:
\begin{equation}
r = \frac{\sum_{i=1}^{n} (x_i - \bar{x})(y_i - \bar{y})}{\sqrt{\sum_{i=1}^{n}(x_i - \bar{x})^2 \cdot \sum_{i=1}^{n}(y_i - \bar{y})^2}}
\end{equation}

\noindent where $\bar{x}$ and $\bar{y}$ are the sample means. We complement Pearson's $r$ with the Mean Absolute Error (MAE), which quantifies the average absolute difference between the scores produced by two benchmarks:
\begin{equation}
\text{MAE} = \frac{1}{n} \sum_{i=1}^{n} \left| y_i - x_i \right|
\end{equation}

Even if a synthetic benchmark is not expected to replicate absolute accuracy scores, it must at minimum preserve the relative ordering of models: a model that ranks higher on real data should also rank higher on synthetic data. 
We measure this property with Spearman's rank correlation, computed from the rank differences $d_i$ (difference between the
positions, based on a sorted evaluation metric, assigned to the same model by two benchmarks) across $n$ models:
\begin{equation}
\rho = 1 - \frac{6 \sum_{i=1}^{n} d_i^2}{n(n^2 - 1)}
\end{equation}

Both $r$ and $\rho$ range from $-1$ (perfect negative alignment) to $+1$ (perfect positive alignment). 
Following established guidelines~\cite{granato2014statistics}, we interpret intermediate values as weak (0--0.5), moderate (0.5--0.75), semi-strong (0.75--0.85), or strong (0.85--1), while noting that these thresholds are not universally fixed.

We also examine the distributional properties that may explain benchmarking performance, following prior analyses~\cite{borsukiewicz2025beyond, boutros2023idiff}. 
To quantify the domain gap (statistical difference) between synthetic and real images, we use the Fr\'{e}chet Inception Distance (FID)~\cite{heusel2017gans}, formulated as:
\begin{equation}
\text{FID} = \|\mu_r - \mu_g\|_2^2 + \text{Tr}\!\left(\Sigma_r + \Sigma_g - 2\left(\Sigma_r \Sigma_g\right)^{1/2}\right) 
\end{equation}

\noindent where the subscripts $r$ and $g$ denote the real and generated distributions, respectively. The mean term measures distributions center shift, while the trace term evaluates mismatches in the shape and orientation of their covariance structures.
To establish statistical significance, we report 95\% confidence intervals obtained by bootstrapping (10{,}000 resamples with replacement), taking the 2.5th and 97.5th percentiles as interval lower and upper bounds.

\subsection{Evaluation Protocols}
We followed the evaluation protocol defined for each real dataset to preserve methodological consistency. 
LFW, CPLFW, CALFW, and AgeDB-30 all use the \textit{Unrestricted with Labeled Outside Data} procedure, in which 3{,}000 mated pairs and 3{,}000 non-mated pairs are divided into 10 folds. 
In each fold, an optimal threshold is computed and then applied to compute accuracy; final results are the mean accuracy $\pm$ standard deviation across folds. CFP-FP follows the same procedure but increases the sample size to 3{,}500 mated and 3{,}500 non-mated pairs.

IJB-B and IJB-C use a template-based protocol in which multiple images or frames are aggregated per identity to approximate real-world conditions. We use the \textit{1:1 Baseline Verification} protocol with 8{,}010{,}270 comparisons to compute TAR at FARs of 1\% and 0.01\%.

For the synthetic datasets, we adopted the \textit{Unrestricted with Labeled Outside Data} protocol on a randomly selected set of 3{,}000 mated and 3{,}000 non-mated pairs, matching the majority of the real dataset protocols. 
The image-pair lists are included in our replication package.

In addition to the protocol-specified metrics, we computed EER, FMR100, FMR1000, and ZeroFMR for every model-dataset combination. 
This provides a more complete picture of how synthetic benchmarks behave across different security and usability requirements.

\begin{figure*}[h!]
    \centering
    \includegraphics[width=1\linewidth]{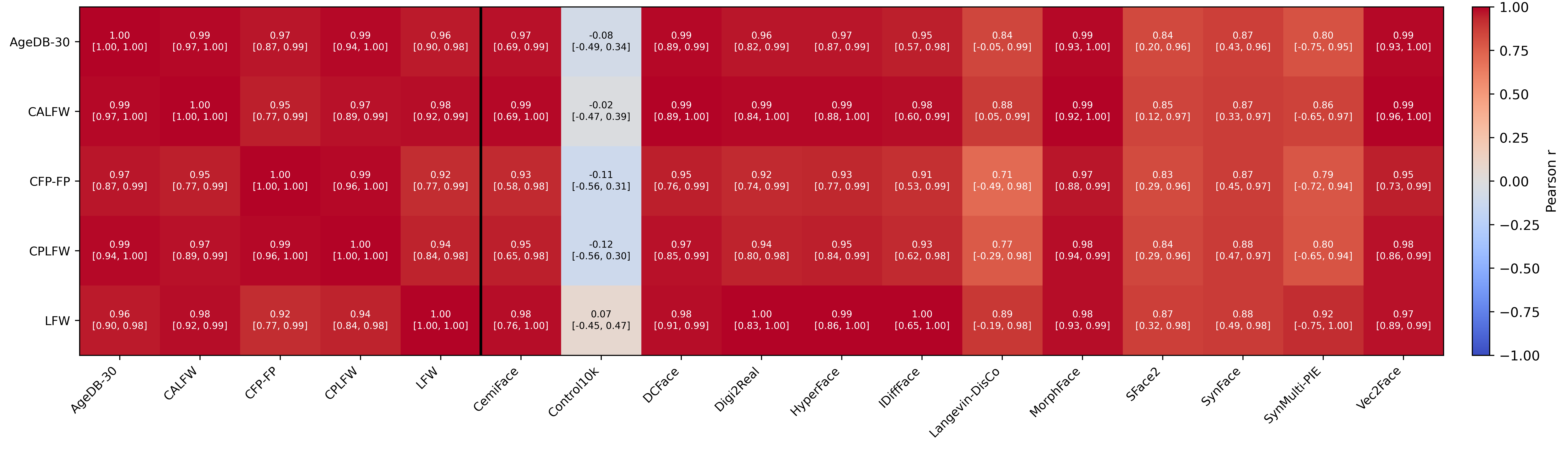}
    \caption{Recognition accuracy alignment between datasets across 24 models - Pearson's $r$ with 95\% CI}
    \label{fig:acc_pearson}
\end{figure*} 
\section{Experiments}

For every model-dataset pair, we first compute cosine similarities and the resulting recognition accuracies. 
We then evaluate biometric performance metrics (EER, FMR100, FMR1000, and ZeroFMR) and measure Pearson's $r$ and Spearman's $\rho$ between datasets. IJB-B and IJB-C were excluded from the accuracy computation as the template-based protocol is incompatible with the pair-based synthetic setup.
To shed light on the factors behind benchmarking quality, we also examine similarity score distributions and FID scores. 
We close with a focused analysis of the two strongest synthetic candidates.

\subsection{Pearson's Correlation}
Recognition accuracies assessed with Pearson's $r$ (Figure~\ref{fig:acc_pearson}) reveal that most dataset pairs exhibit high correlations, with values rarely falling below 0.9. 
The one clear outlier is ControlFace10k, whose low $r$ indicates that it does not satisfy the criteria for a reliable benchmark. 
Analysis of 95\% confidence intervals, however, exposes additional weaknesses that the point estimates conceal. 
Real-real pairs maintain consistently high lower bounds (0.77--0.97), whereas several real-synthetic pairs show substantially larger drops, with Langevin-DisCo and \mbox{SynMulti-PIE} reaching negative lower bounds. 
DCFace, Digi2Real, HyperFace, MorphFace, and Vec2Face, by contrast, maintain confidence bounds comparable to those of real-real pairs.

\begin{table}[ht]
    \small
    \centering
    \caption{Average Pearson's $r$ per metric on 24 models across synthetic datasets when compared with 7 real datasets. Best in bold.}
    \vspace{2mm}
    \label{tab:pearson}
    \setlength{\tabcolsep}{2pt}
    \begin{tabular}{lccccc}
    \toprule
    \textbf{Dataset} & \textbf{EER} & \textbf{ZeroFMR} & \textbf{FMR100} & \textbf{FMR1000} & \textbf{Mean} \\
    \midrule
    CemiFace        & \phantom{-}0.97 & \phantom{-}0.72 & \phantom{-}0.88 & \phantom{-}0.89 & \phantom{-}0.87 \\
    ControlFace10k  &            -0.31 & -0.44           & -0.57           & -0.54           & -0.46           \\
    DCFace          & \phantom{-}0.98 & \phantom{-}0.78 & \phantom{-}0.89 & \phantom{-}0.91 & \phantom{-}0.89 \\
    Digi2Real       & \phantom{-}0.97 & \phantom{-}0.62 & \phantom{-}0.88 & \phantom{-}0.78 & \phantom{-}0.81 \\
    HyperFace       & \phantom{-}0.97 & \phantom{-}0.65 & \phantom{-}0.89 & \phantom{-}0.80 & \phantom{-}0.83 \\
    IDiff-Face      & \phantom{-}0.96 & \phantom{-}0.64 & \phantom{-}0.90 & \phantom{-}0.85 & \phantom{-}0.84 \\
    Langevin-DisCo  & \phantom{-}0.81 & \phantom{-}0.07 & \phantom{-}0.53 & \phantom{-}0.19 & \phantom{-}0.40 \\
    MorphFace       & \textbf{\phantom{-}0.98} & \phantom{-}0.78 &  \textbf{\phantom{-}0.93} &  \textbf{\phantom{-}0.92} &  \textbf{\phantom{-}0.90} \\
    SFace2          & \phantom{-}0.84 & \phantom{-}0.30 & \phantom{-}0.78 & \phantom{-}0.51 & \phantom{-}0.61 \\
    SynFace         & \phantom{-}0.87 & \phantom{-}0.52 & \phantom{-}0.83 & \phantom{-}0.73 & \phantom{-}0.74 \\
    SynMulti-PIE    & \phantom{-}0.85 & \phantom{-}0.35 & \phantom{-}0.78 & \phantom{-}0.59 & \phantom{-}0.64 \\
    Vec2Face        & \phantom{-}0.98 &  \textbf{\phantom{-}0.81} & \phantom{-}0.90 & \phantom{-}0.92 & \phantom{-}0.90 \\
    \bottomrule
    \end{tabular}
\end{table}
A finer-grained view by biometric metric (Table~\ref{tab:pearson}) shows that synthetic datasets achieve their strongest correlations under EER and their weakest under ZeroFMR. 
The ZeroFMR gap points to an unnatural distribution of outliers, specifically highly similar non-mated pairs as documented by Borsukiewicz et al.~\cite{borsukiewicz2025beyond}, even though overall separability remains fairly consistent. 
In line with the accuracy results, MorphFace and Vec2Face rank as the best benchmarks, both averaging $r = 0.90$. 
The two are interchangeable for EER and FMR1000; MorphFace leads on FMR100 (0.93 vs.\ 0.90) while Vec2Face leads on ZeroFMR (0.81 vs.\ 0.78). 
The datasets designed explicitly for evaluation, ControlFace10k and SynMulti-PIE, are among the worst performers, with ControlFace10k being the only dataset that consistently shows negative correlations.

\subsection{Spearman's Rank Correlation}
\begin{figure*}[h]
    \centering
    \includegraphics[width=1\linewidth]{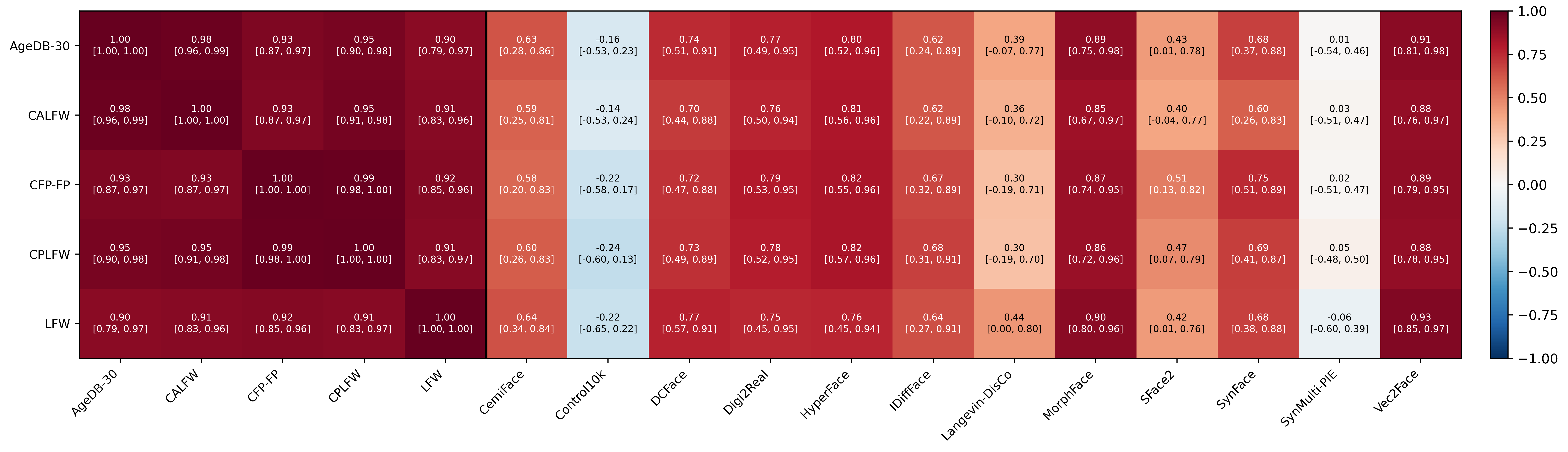}
    \caption{Spearman's $\rho$ with 95\% CI represents how strongly 24 models' accuracies ranking is preserved across datasets.}
    \label{fig:acc_spearman}
\end{figure*}

Pearson's $r$ assumes a linear relationship, which may not hold. 
Spearman's $\rho$ (Figure~\ref{fig:acc_spearman}) relaxes this assumption and directly measures how well model rankings are preserved across datasets. 
The results draw a clear line between synthetic datasets that can rank models and those that cannot. 
MorphFace and Vec2Face show the strongest alignment with real datasets ($\rho > 0.85$), only slightly below the agreement among real-real pairs, confirming them as the most suitable synthetic alternatives. 
The synthetic datasets that perform well do so consistently across all seven real benchmarks, including those focused on extreme poses or large age gaps, while poor synthetic benchmarks underperform just as uniformly.

Spearman's $\rho$ for biometric metrics (Table~\ref{tab:spearman}) is broadly consistent with the Pearson results. 
MorphFace and Vec2Face again lead with a mean $\rho$ of 0.87, while ControlFace10k ($-0.52$) and SynMulti-PIE (0.03) fall far behind. 
Most remaining datasets produce $\rho$ values between 0.5 and 0.8, indicating a moderate-to-strong relationship that supports the feasibility of synthetic benchmarking for face recognition.

\begin{table}[ht]
    \small
    \centering
    \caption{Avg. Spearman's $\rho$ per metric on 24 models across synthetic datasets when compared with 7 real datasets. Best in bold.}
    \vspace{2mm}
    \label{tab:spearman}
    \setlength{\tabcolsep}{2pt}
    \begin{tabular}{lccccc}
    \toprule
    \textbf{Dataset} & \textbf{EER} & \textbf{ZeroFMR} & \textbf{FMR100} & \textbf{FMR1000} & \textbf{Mean} \\
    \midrule
    CemiFace        & \phantom{-}0.65 & \phantom{-}0.59 & \phantom{-}0.64 & \phantom{-}0.71 & \phantom{-}0.65 \\
    ControlFace10k  &            -0.38 &            -0.56 &            -0.58 &            -0.55 &            -0.52 \\
    DCFace          & \phantom{-}0.77 & \phantom{-}0.67 & \phantom{-}0.78 & \phantom{-}0.86 & \phantom{-}0.77 \\
    Digi2Real       & \phantom{-}0.78 & \phantom{-}0.65 & \phantom{-}0.78 & \phantom{-}0.69 & \phantom{-}0.73 \\
    HyperFace       & \phantom{-}0.79 & \phantom{-}0.60 & \phantom{-}0.80 & \phantom{-}0.73 & \phantom{-}0.73 \\
    IDiff-Face      & \phantom{-}0.67 & \phantom{-}0.43 & \phantom{-}0.66 & \phantom{-}0.59 & \phantom{-}0.59 \\
    Langevin-DisCo  & \phantom{-}0.39 & \phantom{-}0.29 & \phantom{-}0.54 & \phantom{-}0.36 & \phantom{-}0.40 \\
    MorphFace       & \phantom{-}0.89 &  \textbf{\phantom{-}0.79} &  \textbf{\phantom{-}0.91} & \phantom{-}0.88 &  \textbf{\phantom{-}0.87} \\
    SFace2          & \phantom{-}0.47 & \phantom{-}0.49 & \phantom{-}0.52 & \phantom{-}0.43 & \phantom{-}0.48 \\
    SynFace         & \phantom{-}0.63 & \phantom{-}0.49 & \phantom{-}0.70 & \phantom{-}0.63 & \phantom{-}0.61 \\
    SynMulti-PIE    &            -0.10 & \phantom{-}0.09 & \phantom{-}0.08 & \phantom{-}0.04 & \phantom{-}0.03 \\
    Vec2Face        &  \textbf{\phantom{-}0.89} & \phantom{-}0.77 & \phantom{-}0.90 &  \textbf{\phantom{-}0.90} & \phantom{-}0.87 \\
    \bottomrule
    \end{tabular}
\end{table}

\begin{figure*}[t]
    \centering
    \includegraphics[width=1\linewidth]{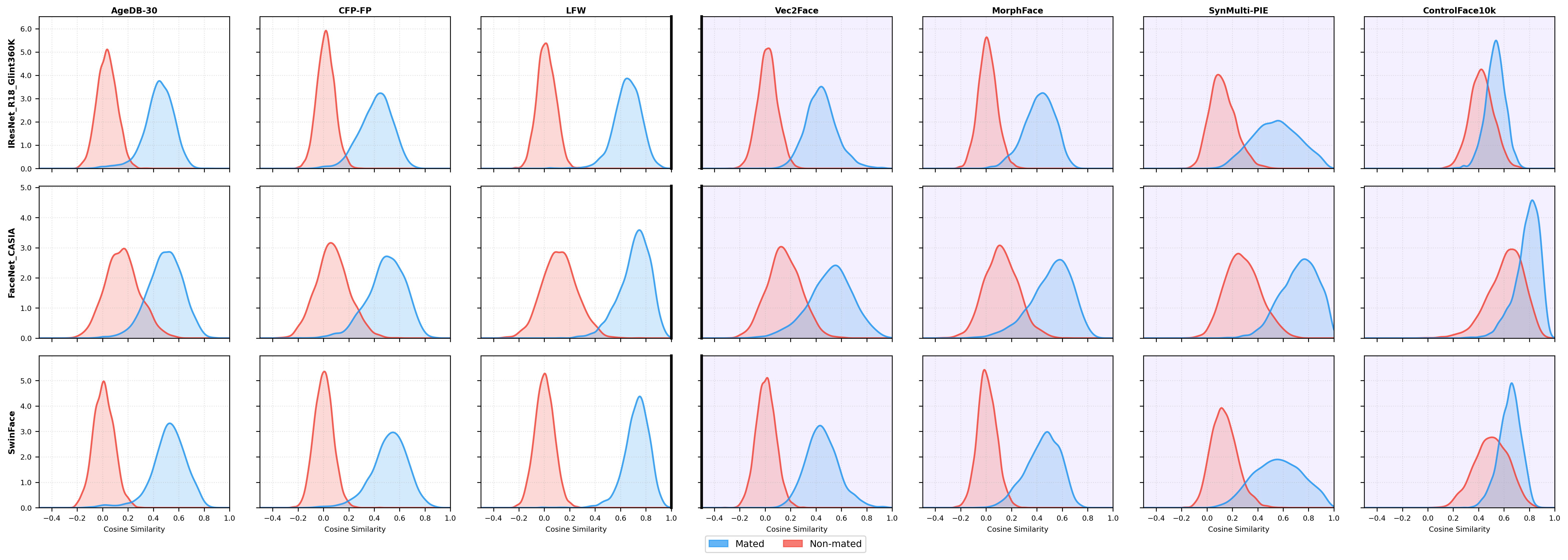}
    \caption{Mated vs. non-mated similarity distributions comparison across individual models and datasets.}
    \label{fig:distr_comp2}
\end{figure*}
\subsection{Fr\'{e}chet Inception Distance}

To quantify the domain gap, we computed FID (Table~\ref{tab:fid_scores}) using an IResNet18 model trained for face recognition, rather than the commonly used Inception-v3~\cite{szegedy2016rethinking}. 
Among real datasets, the highest alignment appears between LFW, CPLFW, and CALFW (FID at most 24.0), which share a substantial portion of their images. 
AgeDB-30 is the most distinct from the other real datasets, likely because of its high age variance, with FID scores in the range 52.1--73.2.

Comparing synthetic to real data reveals that a low FID is necessary but not sufficient for good benchmarking performance. 
Datasets with extremely high FID scores (ControlFace10k, SynFace, SynMulti-PIE, SFace2) consistently show poor correlation with real benchmarks, confirming that some baseline similarity to real data is required. 
Among the top-performing benchmarks, however, FID does not rank them correctly. MorphFace (FID 31.6--64.2) and Vec2Face (FID 51.9--92.0) both outperform CemiFace (FID 26.1--57.9) as benchmarks despite having higher FID scores, which indicates that distributional similarity alone does not determine benchmark reliability.

\begin{table}[h]
    \centering
    \small
    \caption{Cross-dataset FID scores}
    \vspace{2mm}
    \setlength{\tabcolsep}{2pt}
    \label{tab:fid_scores}
    \begin{tabular}{l@{\hspace{-2pt}}ccccc}
        \toprule
        \textbf{Dataset} & \textbf{AgeDB-30} & \textbf{CALFW} & \textbf{CPLFW} & \textbf{CFP-FP} & \textbf{LFW} \\
        \midrule
        AgeDB-30       & --  & \phantom{0}52.1  & \phantom{0}63.4  & \phantom{0}73.2  & \phantom{0}62.2  \\
        CALFW          & \phantom{0}52.1             & --  & \phantom{0}24.0  & \phantom{0}44.6  & \phantom{0}11.1  \\
        CPLFW          & \phantom{0}63.4             & \phantom{0}24.0  & --  & \phantom{0}29.3  & \phantom{0}22.3  \\
        CFP-FP         & \phantom{0}73.2             & \phantom{0}44.6  & \phantom{0}29.3  & --  & \phantom{0}47.2  \\
        LFW            & \phantom{0}62.2             & \phantom{0}11.1  & \phantom{0}22.3  & \phantom{0}47.2  & --  \\
        \midrule
        CemiFace       & \phantom{0}57.9  & \phantom{0}30.6  & \phantom{0}26.1  & \phantom{0}31.5  & \phantom{0}32.9  \\
        ControlFace10k & 337.3            & 346.7            & 237.6            & 230.0            & 351.5            \\
        DCFace         & \phantom{0}67.6  & \phantom{0}41.5  & \phantom{0}36.6  & \phantom{0}39.6  & \phantom{0}44.9  \\
        Digi2Real      & 137.4            & \phantom{0}93.0  & 118.8            & 120.5            & \phantom{0}96.3  \\
        HyperFace      & \phantom{0}98.5  & \phantom{0}57.2  & \phantom{0}70.8  & \phantom{0}78.7  & \phantom{0}58.1  \\
        IDiff-Face     & \phantom{0}89.4  & \phantom{0}58.6  & \phantom{0}57.4  & \phantom{0}63.7  & \phantom{0}61.2  \\
        Langevin-DisCo & 111.8            & \phantom{0}85.4  & \phantom{0}60.9  & \phantom{0}65.4  & \phantom{0}87.1  \\
        MorphFace      & \phantom{0}64.2  & \phantom{0}33.5  & \phantom{0}31.6  & \phantom{0}36.6  & \phantom{0}35.5  \\
        SFace2         & 176.0            & 171.5            & 108.4            & 107.4            & 174.8            \\
        SynFace        & 196.6            & 181.2            & 126.3            & 125.4            & 182.2            \\
        SynMulti-PIE   & 182.2            & 157.6            & 122.7            & 121.7            & 158.7            \\
        Vec2Face       & \phantom{0}92.0  & \phantom{0}51.9  & \phantom{0}68.3  & \phantom{0}78.6  & \phantom{0}53.5  \\
        \bottomrule
    \end{tabular}
\end{table}

\subsection{Similarity Distributions}

To understand why some synthetic datasets succeed and others fail, we examined the cosine similarity distributions for mated and non-mated pairs (Figure~\ref{fig:distr_comp2}). 
On real datasets, SwinFace and IResNet18 produce similar performance levels while FaceNet\_CASIA clearly underperforms; Vec2Face and MorphFace reproduce this ordering, but SynMulti-PIE and ControlFace10k do not.

The shape of the distributions is also informative. 
Vec2Face and MorphFace resemble AgeDB-30 and CFP-FP, suggesting that they introduce more variability than LFW and therefore pose a greater challenge to models. 
This property makes them well suited for evaluating robustness to pose and age variation. 
SynMulti-PIE, by contrast, shows nearly identical overlap between mated and non-mated distributions regardless of model quality, indicating that some of its pairs are too diverse for any reasonable model to match, while others lack the diversity needed to challenge recognizers. 
ControlFace10k suffers from the opposite problem. 
High similarity scores among non-mated pairs signal low between-class variability and poor identity separability.

\subsection{Analysis of MorphFace and Vec2Face}
Given the results above, we focus on MorphFace and Vec2Face to assess their suitability as benchmarks more closely.

A per-dataset Spearman's $\rho$ comparison (Table~\ref{tab:comp_with_real_rho}) shows that MorphFace and Vec2Face are often more aligned with real benchmarks than the real benchmarks are with one another. 
This is most visible for ZeroFMR, where \mbox{MorphFace} ($\rho = 0.79$) exceeds all real-real pairs (0.58--0.78) and Vec2Face (0.77) is close behind. 
The generally lower $\rho$ values for ZeroFMR across the board confirm that measuring performance at zero false positives is inherently more variable, even among real benchmarks. 
The minimum 95\% CI bounds show that alignment between individual dataset pairs can drop lower, but this is expected because certain real benchmarks serve specialized roles such as evaluating robustness to extreme pose or age differences. 
Overall, the evaluated metrics show no systematic gap between real and synthetic data for benchmarking purposes.

\begin{figure}[h]
    \centering
    \includegraphics[width=1.\linewidth]{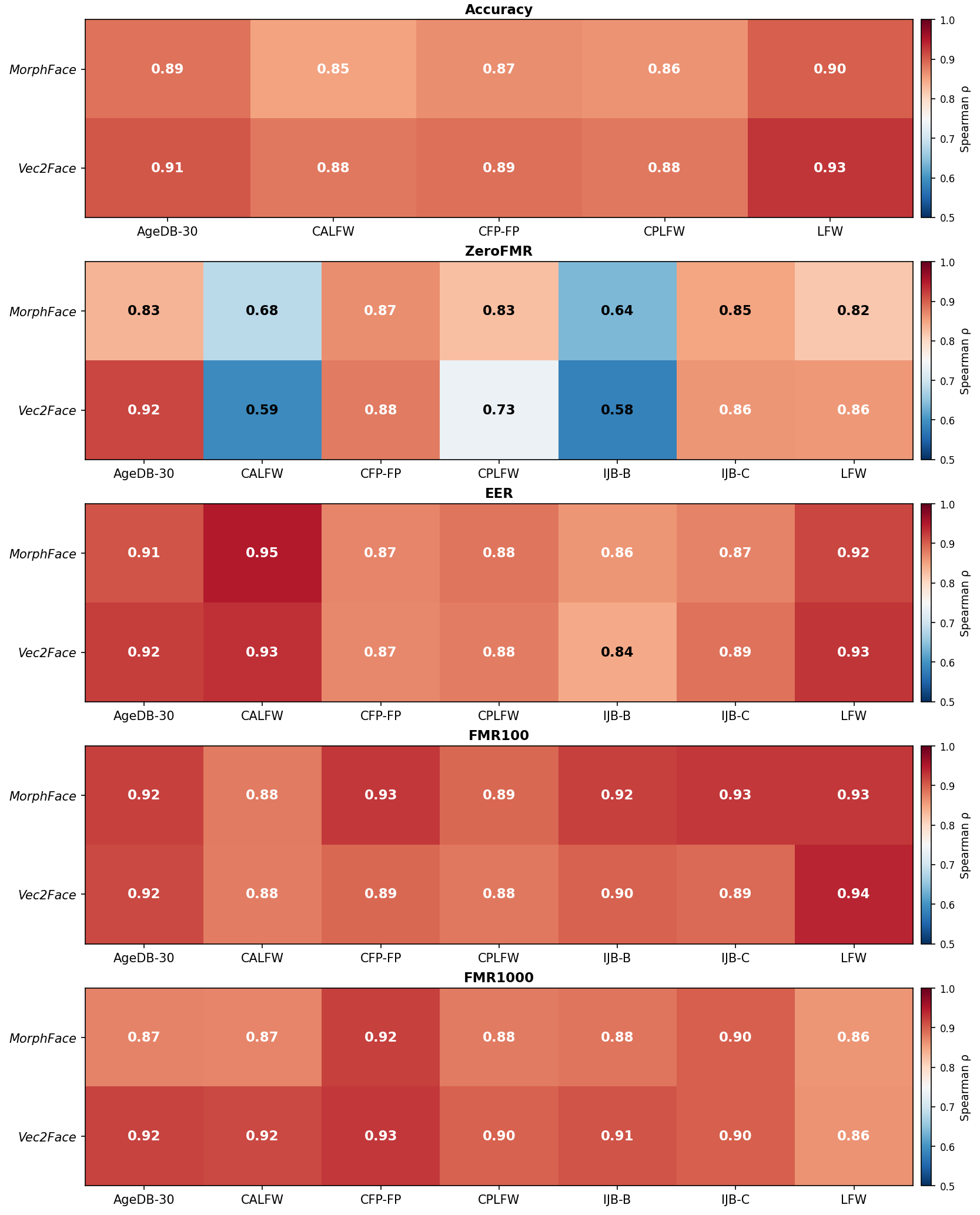}
    \vspace{0mm}
    \caption{Per-metric Spearman's $\rho$ on 24 models -- MorphFace and Vec2Face vs. real datasets.}
    \label{fig:morph_vec}
\end{figure}

Pairwise comparisons (Figure~\ref{fig:morph_vec}) confirm that alignment levels are broadly consistent across metrics. 
\mbox{MorphFace} and Vec2Face position themselves as the most reliable proxies of LFW for recognition accuracy and FMR100, of CFP-FP for FMR1000, and of CALFW for EER. 
The only divergence appears for ZeroFMR, where Vec2Face tracks AgeDB-30 most closely while MorphFace corresponds best to CFP-FP. 
Comparing the two datasets head to head, Vec2Face has a slight edge for accuracy and FMR1000, while MorphFace is marginally better for FMR100. Differences for ZeroFMR and EER are inconsistent.

\begin{table}[h]
\centering
\caption{Per-metric and per-dataset Spearman's $\rho$ coefficient on 24 models. Breakdown with respect to real datasets (excluding self-comparisons). \mbox{MinLowCI} represents the minimum observed value of the lower CI bound.}
\vspace{2mm}
\label{tab:comp_with_real_rho}
\setlength{\tabcolsep}{1pt}
\footnotesize
\begin{tabular}{lcccccccccc}
\toprule
\textbf{Metric} & 
    & \rotatebox{90}{\textbf{LFW}} 
    & \rotatebox{90}{\textbf{AgeDB-30}} 
    & \rotatebox{90}{\textbf{CALFW}} 
    & \rotatebox{90}{\textbf{CFP-FP}} 
    & \rotatebox{90}{\textbf{CPLFW}} 
    & \rotatebox{90}{\textbf{IJB-B}} 
    & \rotatebox{90}{\textbf{IJB-C}} 
    & \rotatebox{90}{\textbf{MorphFace}} 
    & \rotatebox{90}{\textbf{Vec2Face}} \\
\midrule
\multirow{2}{*}{EER}
    & Mean $\rho$ & \phantom{-}0.93 & \phantom{-}0.94 & \phantom{-}0.89 & \phantom{-}0.94 & \phantom{-}0.93 & \phantom{-}0.94 & \phantom{-}0.95 & \phantom{-}0.89 & \phantom{-}0.89 \\
    & MinLowCI    & \phantom{-}0.80 & \phantom{-}0.75 & \phantom{-}0.65 & \phantom{-}0.81 & \phantom{-}0.80 & \phantom{-}0.80 & \phantom{-}0.88 & \phantom{-}0.70 & \phantom{-}0.67 \\
\midrule
\multirow{2}{*}{ZeroFMR}
    & Mean $\rho$ & \phantom{-}0.78 & \phantom{-}0.78 & \phantom{-}0.58 & \phantom{-}0.77 & \phantom{-}0.76 & \phantom{-}0.66 & \phantom{-}0.74 & \phantom{-}0.79 & \phantom{-}0.77 \\
    & MinLowCI    & \phantom{-}0.29 & \phantom{-}0.26 & \phantom{-}0.08 & \phantom{-}0.41 & \phantom{-}0.42 & \phantom{-}0.29 & \phantom{-}0.48 & \phantom{-}0.31 & \phantom{-}0.23 \\
\midrule
\multirow{2}{*}{FMR100}
    & Mean $\rho$ & \phantom{-}0.91 & \phantom{-}0.81 & \phantom{-}0.73 & \phantom{-}0.86 & \phantom{-}0.86 & \phantom{-}0.82 & \phantom{-}0.88 & \phantom{-}0.83 & \phantom{-}0.80 \\
    & MinLowCI    & \phantom{-}0.77 & \phantom{-}0.87 & \phantom{-}0.79 & \phantom{-}0.89 & \phantom{-}0.88 & \phantom{-}0.87 & \phantom{-}0.77 & \phantom{-}0.81 & \phantom{-}0.78 \\
\midrule
\multirow{2}{*}{FMR1000}
    & Mean $\rho$ & \phantom{-}0.89 & \phantom{-}0.96 & \phantom{-}0.95 & \phantom{-}0.97 & \phantom{-}0.96 & \phantom{-}0.95 & \phantom{-}0.95 & \phantom{-}0.95 & \phantom{-}0.95 \\
    & MinLowCI    & \phantom{-}0.70 & \phantom{-}0.76 & \phantom{-}0.72 & \phantom{-}0.79 & \phantom{-}0.74 & \phantom{-}0.80 & \phantom{-}0.79 & \phantom{-}0.70 & \phantom{-}0.70 \\
\bottomrule
\end{tabular}
\end{table}

MAE estimates against real benchmarks (Table~\ref{tab:mae}) show that MorphFace and Vec2Face align best on accuracy and EER and diverge more on FMR-based metrics. 
This pattern suggests that the two synthetic datasets serve well as direct accuracy proxies for the first two metrics and as ranking tools for the rest. 
The lowest error values appear most frequently against CALFW and AgeDB-30, indicating that MorphFace and Vec2Face are particularly well suited for evaluating models on age-invariant identity features. 
The two datasets are nearly indistinguishable from each other in MAE, confirming that either can serve as a benchmark.


\begin{table}[h]
    \centering
    \footnotesize
    \caption{MAE of MorphFace (M) and Vec2Face (V) vs. real datasets across 24 models.}
    \label{tab:mae}
    \vspace{2mm}
    \setlength{\tabcolsep}{3pt}
    \begin{tabular}{llcccccccc}
        \toprule
        \textbf{Metric} & 
            & \rotatebox{90}{\textbf{LFW}} 
            & \rotatebox{90}{\textbf{CALFW}} 
            & \rotatebox{90}{\textbf{CFP-FP}} 
            & \rotatebox{90}{\textbf{CPLFW}} 
            & \rotatebox{90}{\textbf{AgeDB-30}} 
            & \rotatebox{90}{\textbf{IJB-B}} 
            & \rotatebox{90}{\textbf{IJB-C}} 
            & \rotatebox{90}{\textbf{Mean}} \\
        \midrule
       \multirow{2}{*}{Accuracy} & M & 0.045 & 0.015 & 0.024 & 0.051 & 0.015 & --    & --    & 0.030 \\
                 & V & 0.038 & 0.020 & 0.028 & 0.056 & 0.018 & --    & --    & 0.032 \\
        \midrule
        \multirow{2}{*}{EER} & M & 0.048 & 0.024 & 0.024 & 0.063 & 0.016 & 0.035 & 0.039 & 0.036 \\
                 & V & 0.041 & 0.031 & 0.031 & 0.069 & 0.020 & 0.029 & 0.031 & 0.036 \\
        \midrule
        \multirow{2}{*}{ZeroFMR} & M & 0.191 & 0.190 & 0.096 & 0.255 & 0.118 & 0.551 & 0.267 & 0.238 \\
                 & V & 0.182 & 0.189 & 0.097 & 0.272 & 0.090 & 0.560 & 0.276 & 0.238 \\
        \midrule
        \multirow{2}{*}{FMR100} & M & 0.111 & 0.039 & 0.057 & 0.147 & 0.059 & 0.101 & 0.104 & 0.088 \\
                 & V & 0.130 & 0.072 & 0.091 & 0.128 & 0.076 & 0.084 & 0.094 & 0.096 \\
        \midrule
         \multirow{2}{*}{FMR1000} & M & 0.195 & 0.094 & 0.063 & 0.136 & 0.099 & 0.131 & 0.141 & 0.123 \\
                 & V & 0.175 & 0.064 & 0.075 & 0.156 & 0.067 & 0.111 & 0.121 & 0.110 \\
        \bottomrule
    \end{tabular}
\end{table}
\section{Discussion}
Our experiments distinguish reliable from unreliable synthetic face datasets for the purpose of model benchmarking. Building on these results, we discuss the factors that determine how well synthetic data aligns with real samples and propose criteria for validating future synthetic datasets.

\subsection{Factors Affecting Benchmarking Quality}
The benchmarking quality of a synthetic dataset appears to be governed primarily by data diversity. The similarity distributions and FID scores support this interpretation. 
MorphFace and Vec2Face, the two strongest synthetic benchmarks, exhibit well-separated mated and non-mated distributions and relatively low FID scores, and they achieve the highest agreement with real benchmarks. Weaker candidates such as Langevin-DisCo, SFace2, and especially ControlFace10k show limited separation~\cite{borsukiewicz2025beyond, melzi2023gandiffface, yeung2024variface}, larger distributional distance from real data, and correspondingly poor benchmarking performance. 
ControlFace10k represents the extreme case of domain gap, combining near-absent identity separability and a large distribution gap, as exemplified by FID score, with the weakest correlations observed in this study.

A related observation connects benchmarking quality to training utility. The synthetic datasets that perform best as benchmarks also tend to perform best for training~\cite{borsukiewicz2025beyond, mi2025morphFace, wu2024vec2face}. 
MorphFace and Vec2Face rank among the strongest candidates in both settings, while Langevin-DisCo and SFace2 remain among the weakest. 
This overlap suggests that the dataset properties supporting effective training, namely high intra-class diversity and clear identity separability, also contribute to reliable benchmarking. 
The relationship is not strictly one-to-one, however. SynFace, for instance, is relatively weak for training~\cite{yeung2024variface} but comparatively stronger for benchmarking, indicating that the connection between training utility and benchmarking utility deserves further investigation.

\subsection{Criteria for Validating Future Synthetic Face Recognition Datasets}

To confirm future synthetic dataset applicability for benchmarking, in line with our methodology, one should empirically confirm that it aligns with real benchmarks.
Importantly, synthetic datasets should:
\begin{itemize}[nosep]
    \item not exceed the natural disagreement level observed between real datasets
    \item maintain agreement at operating points such as EER/FMR100/FMR1000
\end{itemize}

Our results across 24 models show that, when measuring model accuracy, we should strive for both Pearson's $r$ and Spearman's $\rho$ to exceed 0.9, with lower bounds of the 95\% CIs $>$ 0.75. However, specific values depend on the number and choice of the models to be evaluated. 

\section{Conclusions}

We evaluated whether synthetic face datasets can serve as reliable benchmarks for face recognition. 
The results reveal substantial variability across candidates. Datasets designed explicitly for benchmarking do not necessarily produce the strongest agreement with real benchmarks; several synthetic training datasets prove far more effective in that role. 
Among all candidates, MorphFace and Vec2Face stand out, consistently achieving Pearson's $r$ and Spearman's $\rho$ values whose 95\% confidence intervals overlap with the disagreement observed among real benchmarks themselves.

Beyond identifying the strongest candidates, our results show a consistent separation between high-fidelity and low-fidelity synthetic benchmarks across all biometric metrics, a separation that is also reflected in the underlying similarity distributions. 
Well-constructed synthetic datasets can therefore support not only training but also reliable comparative evaluation for face recognition, reducing reliance on privacy-sensitive real facial data.

A promising direction for future work is the construction of synthetic evaluation pairs under explicit diversity constraints spanning both subject attributes and acquisition conditions. 
Such a strategy could yield stronger alignment with specialized benchmarks and enable targeted evaluation under specific covariates such as pose, age, or occlusion.

\section*{Acknowledgments}
\begin{itemize}[nosep]
    \item This research was funded by the Luxembourg Army.
    \item AI-based tools were used for fixing grammatical and typographical errors.  
\end{itemize}

{\small
\bibliographystyle{ieee}
\bibliography{main}

@article{borsukiewicz2025beyond,
  title={Beyond real faces: synthetic datasets can achieve reliable recognition performance without privacy compromise},
  author={Borsukiewicz, Pawe{\l} and Boutros, Fadi and Olatunji, Iyiola E and Beumier, Charles and Ou{\'e}draogo, Wendk{\^u}uni C and Klein, Jacques and Bissyand{\'e}, Tegawend{\'e} F},
  journal={npj Artificial Intelligence},
  year={2026},
  publisher={Nature Publishing Group}
}

@article{nisevic2025synthetic,
  title={Synthetic data in medicine: Legal and ethical considerations for patient profiling},
  author={Nisevic, Maja and Milojevic, Dusko and Spajic, Daniela},
  journal={Computational and Structural Biotechnology Journal},
  volume={28},
  pages={190--198},
  year={2025},
  publisher={Elsevier}
}

@article{boutros2023synthetic,
  title={Synthetic data for face recognition: Current state and future prospects},
  author={Boutros, Fadi and Struc, Vitomir and Fierrez, Julian and Damer, Naser},
  journal={Image and Vision Computing},
  volume={135},
  pages={104688},
  year={2023},
  publisher={Elsevier}
}

@article{sun2025ensuring,
  title={Ensuring privacy in face recognition: a survey on data generation, inference and storage},
  author={Sun, Zhifang and Liu, Zhe},
  journal={Discover Applied Sciences},
  volume={7},
  number={5},
  pages={441},
  year={2025},
  publisher={Springer}
}

@online{GDPR2016a,
  date       = {2016-05-04},
  location   = {OJ L 119, 4.5.2016, p. 1--88},
  title      = {Regulation ({EU}) 2016/679 of the {European} {Parliament} and of the {Council}},
  url        = {https://data.europa.eu/eli/reg/2016/679/oj},
  titleaddon = {of 27 {April} 2016 on the protection of natural persons with regard to the processing of personal data and on the free movement of such data, and repealing {Directive} 95/46/{EC} ({General} {Data} {Protection} {Regulation})},
  author     = {{European Parliament} and {Council of the European Union}},
  urldate    = {\today},
}

@online{BIPA,
	author = {{Illinois General Assembly}},
	title = {Public Act 095-0994},
	year = {2008},
}

@inproceedings{colbois2021use,
  title={On the use of automatically generated synthetic image datasets for benchmarking face recognition},
  author={Colbois, Laurent and de Freitas Pereira, Tiago and Marcel, S{\'e}bastien},
  booktitle={2021 IEEE International Joint Conference on Biometrics (IJCB)},
  pages={1--8},
  year={2021},
  organization={IEEE}
}

@inproceedings{baltsou2024sdfd,
  title={Sdfd: Building a versatile synthetic face image dataset with diverse attributes},
  author={Baltsou, Georgia and Sarridis, Ioannis and Koutlis, Christos and Papadopoulos, Symeon},
  booktitle={2024 IEEE 18th International Conference on Automatic Face and Gesture Recognition (FG)},
  pages={1--10},
  year={2024},
  organization={IEEE}
}

@inproceedings{nzalasse2024sig,
  title={SIG: A Synthetic Identity Generation Pipeline for Generating Evaluation Datasets for Face Recognition},
  author={Nzalasse, Kassi and Raj, Rishav and Laird, Eli and Clark, Corey},
  booktitle={International Conference on Pattern Recognition},
  pages={299--313},
  year={2024},
  organization={Springer}
}

@inproceedings{huang2008lfw,
  title={Labeled faces in the wild: A database forstudying face recognition in unconstrained environments},
  author={Huang, Gary B and Mattar, Marwan and Berg, Tamara and Learned-Miller, Eric},
  booktitle={Workshop on faces in'Real-Life'Images: detection, alignment, and recognition},
  year={2008}
}

@inproceedings{sengupta2016cfp_fp,
  title={Frontal to profile face verification in the wild},
  author={Sengupta, Soumyadip and Chen, Jun-Cheng and Castillo, Carlos and Patel, Vishal M and Chellappa, Rama and Jacobs, David W},
  booktitle={2016 IEEE winter conference on applications of computer vision (WACV)},
  pages={1--9},
  year={2016},
  publisher={IEEE}
}

@article{zheng2018cplfw,
  title={Cross-pose lfw: A database for studying cross-pose face recognition in unconstrained environments},
  author={Zheng, Tianyue and Deng, Weihong},
  journal={Beijing University of Posts and Telecommunications, Tech. Rep},
  volume={5},
  number={7},
  pages={5},
  year={2018}
}

@inproceedings{moschoglou2017agedb,
  title={Agedb: the first manually collected, in-the-wild age database},
  author={Moschoglou, Stylianos and Papaioannou, Athanasios and Sagonas, Christos and Deng, Jiankang and Kotsia, Irene and Zafeiriou, Stefanos},
  booktitle={proceedings of the IEEE conference on computer vision and pattern recognition workshops},
  pages={51--59},
  year={2017}
}

@article{zheng2017calfw,
  title={Cross-age lfw: A database for studying cross-age face recognition in unconstrained environments},
  author={Zheng, Tianyue and Deng, Weihong and Hu, Jiani},
  journal={arXiv preprint arXiv:1708.08197},
  year={2017}
}

@inproceedings{whitelam2017iarpa,
  title={Iarpa janus benchmark-b face dataset},
  author={Whitelam, Cameron and Taborsky, Emma and Blanton, Austin and Maze, Brianna and Adams, Jocelyn and Miller, Tim and Kalka, Nathan and Jain, Anil K and Duncan, James A and Allen, Kristen and others},
  booktitle={proceedings of the IEEE conference on computer vision and pattern recognition workshops},
  pages={90--98},
  year={2017}
}

@inproceedings{maze2018iarpa,
  title={Iarpa janus benchmark-c: Face dataset and protocol},
  author={Maze, Brianna and Adams, Jocelyn and Duncan, James A and Kalka, Nathan and Miller, Tim and Otto, Charles and Jain, Anil K and Niggel, W Tyler and Anderson, Janet and Cheney, Jordan and others},
  booktitle={2018 international conference on biometrics (ICB)},
  pages={158--165},
  year={2018},
  organization={IEEE}
}

@article{shahreza2024hyperface,
  title={HyperFace: Generating synthetic face recognition datasets by exploring face embedding hypersphere},
  author={Shahreza, Hatef Otroshi and Marcel, S{\'e}bastien},
  journal={arXiv preprint arXiv:2411.08470},
  year={2024}
}

@article{boutros2024sface2,
  title={Sface2: Synthetic-based face recognition with w-space identity-driven sampling},
  author={Boutros, Fadi and Huber, Marco and Luu, Anh Thi and Siebke, Patrick and Damer, Naser},
  journal={IEEE Transactions on Biometrics, Behavior, and Identity Science},
  year={2024},
  publisher={IEEE}
}

@inproceedings{boutros2023idiff,
  title={Idiff-face: Synthetic-based face recognition through fizzy identity-conditioned diffusion model},
  author={Boutros, Fadi and Grebe, Jonas Henry and Kuijper, Arjan and Damer, Naser},
  booktitle={Proceedings of the IEEE/CVF International Conference on Computer Vision},
  pages={19650--19661},
  year={2023}
}

@inproceedings{kim2023dcface,
  title={Dcface: Synthetic face generation with dual condition diffusion model},
  author={Kim, Minchul and Liu, Feng and Jain, Anil and Liu, Xiaoming},
  booktitle={Proceedings of the ieee/cvf conference on computer vision and pattern recognition},
  pages={12715--12725},
  year={2023}
}

@article{sun2024cemiface,
  title={Cemiface: Center-based semi-hard synthetic face generation for face recognition},
  author={Sun, Zhonglin and Song, Siyang and Patras, Ioannis and Tzimiropoulos, Georgios},
  journal={Advances in Neural Information Processing Systems},
  volume={37},
  pages={35612--35638},
  year={2024}
}

@article{wu2024vec2face,
  title={Vec2Face: Scaling face dataset generation with loosely constrained vectors},
  author={Wu, Haiyu and Singh, Jaskirat and Tian, Sicong and Zheng, Liang and Bowyer, Kevin W},
  journal={arXiv preprint arXiv:2409.02979},
  year={2024}
}

@inproceedings{mi2025morphFace,
  title={Data Synthesis with Diverse Styles for Face Recognition via 3DMM-Guided Diffusion},
  author={Mi, Yuxi and Zhong, Zhizhou and Huang, Yuge and Yuan, Qiuyang and Zhao, Xuan and Xu, Jianqing and Ding, Shouhong and Wang, Shaoming and Guo, Rizen and Zhou, Shuigeng},
  booktitle={Proceedings of the Computer Vision and Pattern Recognition Conference},
  pages={21203--21214},
  year={2025}
}

@article{geissbuhler2024Langevin,
  title={Synthetic face datasets generation via latent space exploration from brownian identity diffusion},
  author={Geissb{\"u}hler, David and Shahreza, Hatef Otroshi and Marcel, S{\'e}bastien},
  journal={arXiv preprint arXiv:2405.00228},
  year={2024}
}

@inproceedings{qiu2021synface,
  title={Synface: Face recognition with synthetic data},
  author={Qiu, Haibo and Yu, Baosheng and Gong, Dihong and Li, Zhifeng and Liu, Wei and Tao, Dacheng},
  booktitle={Proceedings of the IEEE/CVF International Conference on Computer Vision},
  pages={10880--10890},
  year={2021}
}

@article{qin2023swinface,
  title={SwinFace: A multi-task transformer for face recognition, expression recognition, age estimation and attribute estimation},
  author={Qin, Lixiong and Wang, Mei and Deng, Chao and Wang, Ke and Chen, Xi and Hu, Jiani and Deng, Weihong},
  journal={IEEE Transactions on Circuits and Systems for Video Technology},
  volume={34},
  number={4},
  pages={2223--2234},
  year={2023},
  publisher={IEEE}
}

@inproceedings{dan2023transface,
  title={Transface: Calibrating transformer training for face recognition from a data-centric perspective},
  author={Dan, Jun and Liu, Yang and Xie, Haoyu and Deng, Jiankang and Xie, Haoran and Xie, Xuansong and Sun, Baigui},
  booktitle={Proceedings of the IEEE/CVF international conference on computer vision},
  pages={20642--20653},
  year={2023}
}

@article{george2024edgeface,
  title={Edgeface: Efficient face recognition model for edge devices},
  author={George, Anjith and Ecabert, Christophe and Shahreza, Hatef Otroshi and Kotwal, Ketan and Marcel, S{\'e}bastien},
  journal={IEEE Transactions on Biometrics, Behavior, and Identity Science},
  volume={6},
  number={2},
  pages={158--168},
  year={2024},
  publisher={IEEE}
}

@INPROCEEDINGS{SzegedyEtAl2015,
  author={Szegedy, Christian and Wei Liu and Yangqing Jia and Sermanet, Pierre and Reed, Scott and Anguelov, Dragomir and Erhan, Dumitru and Vanhoucke, Vincent and Rabinovich, Andrew},
  booktitle={2015 IEEE Conference on Computer Vision and Pattern Recognition (CVPR)}, 
  title={Going deeper with convolutions}, 
  year={2015},
  pages={1-9},
  doi={10.1109/CVPR.2015.7298594}}

@inproceedings{behrmann2019invertible,
  title={Invertible residual networks},
  author={Behrmann, Jens and Grathwohl, Will and Chen, Ricky TQ and Duvenaud, David and Jacobsen, J{\"o}rn-Henrik},
  booktitle={International conference on machine learning},
  pages={573--582},
  year={2019},
  organization={PMLR}
}

@inproceedings{deng2019arcface,
  title={Arcface: Additive angular margin loss for deep face recognition},
  author={Deng, Jiankang and Guo, Jia and Xue, Niannan and Zafeiriou, Stefanos},
  booktitle={Proceedings of the IEEE/CVF conference on computer vision and pattern recognition},
  pages={4690--4699},
  year={2019}
}

@misc{Glint,
      title={Partial FC: Training 10 Million Identities on a Single Machine}, 
      author={Xiang An and Xuhan Zhu and Yang Xiao and Lan Wu and Ming Zhang and Yuan Gao and Bin Qin and Debing Zhang and Ying Fu},
      year={2021},
      eprint={2010.05222},
      archivePrefix={arXiv},
      primaryClass={cs.CV},
      url={https://arxiv.org/abs/2010.05222}, 
}

@INPROCEEDINGS{MS1MV3,
  author={Deng, Jiankang and Guo, Jia and Zhang, Debing and Deng, Yafeng and Lu, Xiangju and Shi, Song},
  booktitle={2019 IEEE/CVF International Conference on Computer Vision Workshop (ICCVW)}, 
  title={Lightweight Face Recognition Challenge}, 
  year={2019},
  volume={},
  number={},
  pages={2638-2646},
  doi={10.1109/ICCVW.2019.00322}}

@article{yi2014casia,
  title={Learning face representation from scratch},
  author={Yi, Dong and Lei, Zhen and Liao, Shengcai and Li, Stan Z},
  journal={arXiv preprint arXiv:1411.7923},
  year={2014}
}

@misc{VggFace2,
      title={VGGFace2: A dataset for recognising faces across pose and age}, 
      author={Qiong Cao and Li Shen and Weidi Xie and Omkar M. Parkhi and Andrew Zisserman},
      year={2018},
      eprint={1710.08092},
      archivePrefix={arXiv},
      primaryClass={cs.CV},
      url={https://arxiv.org/abs/1710.08092}, 
}

@misc{ISO19795,
  title        = {Information Technology -- Biometric Performance Testing and Reporting -- Part 1: Principles and Framework},
  publisher = {International publisher for Standardization and International Electrotechnical Committee},
  author       = {{ISO/IEC JTC 1/SC 37 Biometrics}},
  number       = {ISO/IEC 19795-1:2021},
  year         = {2021},
}

@article{ho2020denoising,
  title={Denoising diffusion probabilistic models},
  author={Ho, Jonathan and Jain, Ajay and Abbeel, Pieter},
  journal={Advances in neural information processing systems},
  volume={33},
  pages={6840--6851},
  year={2020}
}

@article{goodfellow2014generative,
  title={Generative adversarial nets},
  author={Goodfellow, Ian J and Pouget-Abadie, Jean and Mirza, Mehdi and Xu, Bing and Warde-Farley, David and Ozair, Sherjil and Courville, Aaron and Bengio, Yoshua},
  journal={Advances in neural information processing systems},
  volume={27},
  year={2014}
}

@article{smith2022ethical,
  title={The ethical application of biometric facial recognition technology},
  author={Smith, Marcus and Miller, Seumas},
  journal={Ai \& Society},
  volume={37},
  number={1},
  pages={167--175},
  year={2022},
  publisher={Springer}
}

@article{karras2021alias,
  title={Alias-free generative adversarial networks},
  author={Karras, Tero and Aittala, Miika and Laine, Samuli and H{\"a}rk{\"o}nen, Erik and Hellsten, Janne and Lehtinen, Jaakko and Aila, Timo},
  journal={Advances in neural information processing systems},
  volume={34},
  pages={852--863},
  year={2021}
}

@inproceedings{rombach2022high,
  title={High-resolution image synthesis with latent diffusion models},
  author={Rombach, Robin and Blattmann, Andreas and Lorenz, Dominik and Esser, Patrick and Ommer, Bj{\"o}rn},
  booktitle={Proceedings of the IEEE/CVF conference on computer vision and pattern recognition},
  pages={10684--10695},
  year={2022}
}

@inproceedings{boutros2022sface,
  title={Sface: Privacy-friendly and accurate face recognition using synthetic data},
  author={Boutros, Fadi and Huber, Marco and Siebke, Patrick and Rieber, Tim and Damer, Naser},
  booktitle={2022 IEEE International Joint Conference on Biometrics (IJCB)},
  pages={1--11},
  year={2022},
  organization={IEEE}
}

@article{kortylewski2018training,
  title={Training deep face recognition systems with synthetic data},
  author={Kortylewski, Adam and Schneider, Andreas and Gerig, Thomas and Egger, Bernhard and Morel-Forster, Andreas and Vetter, Thomas},
  journal={arXiv preprint arXiv:1802.05891},
  year={2018}
}

@article{yeung2024variface,
  title={Variface: Fair and diverse synthetic dataset generation for face recognition},
  author={Yeung, Michael and Teramoto, Toya and Wu, Songtao and Fujiwara, Tatsuo and Suzuki, Kenji and Kojima, Tamaki},
  journal={arXiv preprint arXiv:2412.06235},
  year={2024}
}

@inproceedings{melzi2023gandiffface,
  title={Gandiffface: Controllable generation of synthetic datasets for face recognition with realistic variations},
  author={Melzi, Pietro and Rathgeb, Christian and Tolosana, Ruben and Vera-Rodriguez, Ruben and Lawatsch, Dominik and Domin, Florian and Schaubert, Maxim},
  booktitle={Proceedings of the IEEE/CVF International Conference on Computer Vision},
  pages={3086--3095},
  year={2023}
}

@techreport{granato2014statistics,
  title={Statistics for stochastic modeling of volume reduction, hydrograph extension, and water-quality treatment by structural stormwater runoff best management practices (BMPs)},
  author={Granato, Gregory E},
  year={2014},
  institution={US Geological Survey}
}

@article{heusel2017gans,
  title={Gans trained by a two time-scale update rule converge to a local nash equilibrium},
  author={Heusel, Martin and Ramsauer, Hubert and Unterthiner, Thomas and Nessler, Bernhard and Hochreiter, Sepp},
  journal={Advances in neural information processing systems},
  volume={30},
  year={2017}
}

@inproceedings{szegedy2016rethinking,
  title={Rethinking the inception architecture for computer vision},
  author={Szegedy, Christian and Vanhoucke, Vincent and Ioffe, Sergey and Shlens, Jon and Wojna, Zbigniew},
  booktitle={Proceedings of the IEEE conference on computer vision and pattern recognition},
  pages={2818--2826},
  year={2016}
}

@book{grother2021face,
  title={Face recognition vendor test (FRVT) Part 7: Identification for paperless travel and immigration},
  author={Grother, Patrick and Grother, Patrick and Hom, Austin and Ngan, Mei and Hanaoka, Kayee},
  year={2021},
  publisher={US Department of Commerce, National Institute of Standards and Technology}
}

@inproceedings{george2025digi2real,
  title={Digi2real: Bridging the realism gap in synthetic data face recognition via foundation models},
  author={George, Anjith and Marcel, S{\'e}bastien},
  booktitle={Proceedings of the Winter Conference on Applications of Computer Vision},
  pages={1469--1478},
  year={2025}
}
}

\end{document}